\documentclass[11pt,a4paper]{article}
\usepackage{times}
\usepackage{latexsym}
\usepackage[hyperref]{emnlp2020}

\usepackage{microtype}
\usepackage{times}
\usepackage{latexsym}
\usepackage{times}
\usepackage{url}
\usepackage{xcolor}
\usepackage{amsmath}
\usepackage{graphicx}
\usepackage{multirow}
\usepackage{bm}
\usepackage{amsfonts}
\usepackage{mathtools}
\usepackage{amsthm}
\usepackage{amssymb}
\usepackage{physics}
\usepackage[utf8]{inputenc}
\usepackage[english]{babel}
\usepackage{algorithm}
\usepackage[noend]{algpseudocode}
\usepackage{balance}

\newtheorem{definition}{Definition}

\newcommand{\realNumber}{\mathbb{R}}

\aclfinalcopy % Uncomment this line for the final submission
\def\aclpaperid{2185} %  Enter the acl Paper ID here
\newcommand{\embedding}{\mathcal{E}}
\newcommand{\vocabulary}{\bm{W}}

\newcommand{\embedDimen}{d}
\newcommand{\textEmbed}{\bm{x}}
\newcommand{\textEmbedSet}{\mathcal{X}}
\newcommand{\wordEmbed}{\underline{x}}
\newcommand{\textLength}{m}
\newcommand{\network}{\mathbf{N}}
\newcommand{\class}{c}
\newcommand{\classes}{C}

\newcommand{\MSR}{\texttt{MSR}}
\newcommand{\dist}{\epsilon}
\newcommand{\ball}{\mathsf{Ball}}

\title{Assessing Robustness of Text Classification through\\Maximal Safe Radius Computation}

\author{Emanuele La Malfa\textsuperscript{\dag} \hspace{0.5cm} Min Wu\textsuperscript{\dag} \hspace{0.5cm} Luca Laurenti\textsuperscript{\dag} \\
\textbf{Benjie Wang\textsuperscript{\dag} \hspace{0.5cm} Anthony Hartshorn\textsuperscript{\S} \hspace{0.5cm} Marta Kwiatkowska\textsuperscript{\dag}} \\
  \textsuperscript{\dag}Department of Computer Science, University of Oxford, United Kingdom\\
  \textsuperscript{\S}Genie AI, London, United Kingdom\\
  \small{\texttt{\{emanuele.lamalfa, min.wu, luca.laurenti, benjie.wang, marta.kwiatkowska\}@cs.ox.ac.uk}}\\
  \small{\texttt{\{anthony.hartshorn\}@genieai.co}}}

\date{}

\begin{document}
\maketitle
\begin{abstract}
Neural network NLP models are vulnerable to small modifications of the input that maintain the original meaning but result in a different prediction.  
In this paper, we focus on robustness of text classification against word substitutions, aiming to provide guarantees that the model prediction does not change if a word is replaced with a plausible alternative, such as a synonym.
As a measure of robustness, we adopt the notion of the maximal safe radius for a given input text, which is the minimum distance in the embedding space to the decision boundary.
Since computing the exact maximal safe radius is not feasible in practice, we instead approximate it by computing a lower and upper bound.
For the upper bound computation, we employ Monte Carlo Tree Search in conjunction with syntactic filtering to analyse the effect of single and multiple word substitutions. The lower bound computation is achieved through an adaptation of the linear bounding techniques implemented in tools CNN-Cert and POPQORN, respectively for convolutional and recurrent network models. We evaluate the methods on sentiment analysis and news classification models for four datasets (IMDB, SST, AG News and NEWS) and a range of embeddings, and provide an analysis of robustness trends.  
We also apply our framework to interpretability analysis and compare it with LIME.
\end{abstract}

\section{Introduction}
\label{intro}
Deep neural networks (DNNs)  have shown great promise in Natural Language Processing (NLP), outperforming  other machine learning techniques in sentiment analysis~\citep{devlinbert:2018}, language translation~\citep{chorowskiattention:2015}, speech recognition~\citep{jia2018transfer} and many other tasks\footnote{See \url{https://paperswithcode.com/area/natural-language-processing} % for a list of papers, models and their performance.
}.
Despite these successes, concerns have been raised about robustness and interpretability of NLP models~\citep{arras2016explaining}. %various concerns still remain. 
It is known that DNNs are vulnerable to adversarial examples, that is, imperceptible perturbations of a test point that cause a prediction error~\citep{goodfellowexplaining:2014}. In NLP this issue manifests itself as a sensitivity of the prediction to  small modifications of the input text (e.g., replacing a word with a synonym).
In this paper we work with DNNs for text  %sentiment
analysis
and, given a text and a word embedding,  consider the problem of quantifying the robustness of the DNN with respect to word substitutions. In particular, we define the \emph{maximal safe radius (MSR)} of a text as the minimum distance (in the embedding space) of the text from the decision boundary, i.e., from the nearest  %modification of the
perturbed text that is classified differently from the original. Unfortunately, computation of the MSR for a neural network is %generally
an NP-hard problem and becomes impractical for real-world networks  %of practical size
\cite{katzreluplex:2017}. As a consequence, we adapt constraint relaxation techniques \cite{weng2018towards,zhangefficient:2018,wong2018provable} developed % for formal analysis of neural networks \cite{} 
to compute a guaranteed lower bound of the MSR for both \emph{convolutional (CNNs)} and \emph{recurrent neural networks (RNNs)}. Furthermore, in order to compute an upper bound for the MSR we adapt the  Monte Carlo Tree Search (MCTS) algorithm \cite{coulommcts:2007} to word embeddings to search for (syntactically and semantically) plausible word substitutions that result in a classification different from the original; the distance to any such perturbed text is an upper bound, albeit possibly loose. We employ our framework to perform an empirical analysis of the robustness trends of sentiment analysis and news classification tasks for a range of embeddings on vanilla CNN and LTSM models.
In particular, we consider the IMDB dataset~\citep{maaslearning:2011}, the Stanford Sentiment Treebank (SST) dataset~\citep{socherrecursive:2013}, the AG News Corpus Dataset ~\citep{zhangcharacter:2015} and the NEWS Dataset~\citep{vitale2012classification}.
We empirically observe that, although generally NLP models are vulnerable to minor perturbations and their robustness degrades with the dimensionality of the embedding, %and show that, although
in some cases we are able to certify the text's classification against any word substitution. 
Furthermore, we show that our framework can be employed for interpretability analysis by computing a saliency measure for each word, which has the advantage of being able to take into account non-linearties of the decision boundary that local approaches such as LIME \cite{lime:2016} cannot handle.

In summary this paper makes the following main contributions:
\begin{itemize}
    \item We develop a framework for quantifying the robustness of NLP models against (single and multiple) word substitutions based on MSR computation.
    \item We adapt existing techniques for approximating the MSR (notably CNN-Cert, POPQORN and MCTS) to word embeddings and semantically and syntactically plausible word substitutions. 
    \item We evaluate vanilla CNN and LSTM sentiment and news classification models on a range of embeddings and datasets, and provide a systematic analysis of the robustness trends and comparison with LIME on interpretability analysis. 
\end{itemize}

\paragraph{Related Work.}
Deep neural networks are known to be vulnerable to adversarial attacks (small perturbations of the network input that result in a misclassification)~\cite{szegedy2014intriguing,biggio2013evasion,biggio2018wild}. The NLP domain has also been shown to suffer from this issue 
%\MK{Also cite Biggio} \MW{Add two. Check the right papers?} 
%Since we work with text classification, we give an overview of adversarial attacks
~\cite{belinkov2018synthetic,ettinger2017towards,gao2018black,jia2017adversarial,liang2017deep,zhang2020adversarial}. The vulnerabilities of NLP models have been exposed via, for example, small character perturbations~\cite{ebrahimi2018hotflip}, syntactically controlled paraphrasing~\cite{iyyer2018adversarial}, targeted keywords attacks~\cite{alzantotgenerating:2018,cheng2018seq2sick}, and exploitation of back-translation systems~\cite{ribeiro2018semantically}.
Formal verification can guarantee that the classification of an input of a neural network is invariant to perturbations of a certain magnitude, which can be established through the concept of the \emph{maximal safe radius}~\cite{wugame:2019} or, dually, \emph{minimum adversarial distortion}~\cite{WZCYSGHD2018}. 
While verification methods based on constraint solving~\cite{katzreluplex:2017,katz2019marabou} and mixed integer programming~\cite{dutta2018output,CNR2017} can provide complete robustness guarantees, in the sense of computing exact bounds, they are expensive and do not scale to real-world networks because the problem itself is NP-hard~\cite{katzreluplex:2017}. To work around this, incomplete approaches, such as search-based methods~\cite{huangsafety:2017,wu2020robustness} or reachability  computation~\cite{ruan2018reachability},  %\MK{cite IJCAI 2018} \MW{Added.}, 
instead compute looser robustness bounds with much greater scalability, albeit relying on the knowledge of non-trivial Lipschitz constants. 
In this work, we exploit approximate, scalable, linear constraint relaxation methods~\cite{weng2018towards,zhangefficient:2018,wong2018provable}, which do not assume Lipschitz continuity. In particular, we adapt the CNN-Cert tool~\cite{boopathy2019cnn} and its recurrent extension POPQORN~\cite{kopopqorn:2019} to compute robustness guarantees for text classification in the NLP domain. We note that NLP robustness has also been addressed using interval bound propagation~\cite{huang2019achieving,jia2019certified}. 

\section{Robustness Quantification of Text Classification against Word Substitutions}
\label{sec:text-classification}
In text classification an algorithm processes a text and associates it to a category. Raw text, i.e., a sequence of words (or similarly sentences or phrases), is converted to a sequence of real-valued vectors through an embedding $\embedding : \vocabulary \to \textEmbedSet \subseteq \realNumber^\embedDimen$, which %is a function  that 
maps each element of a finite set $\vocabulary$ (e.g., a vocabulary) into a vector of real numbers. %In the literature
There are many different ways to build embeddings~\citep{goldbergword2vec:2014,penningtonglove:2014,wallachtopic:2006}, nonetheless their common objective is to capture relations among words. Furthermore, it is also possible to enforce into the embedding syntactic/semantic constraints, a technique commonly known as counter-fitting~\citep{mrkvsiccounter:2016}, which we assess from a robustness perspective in Section~\ref{sec:experiments}. %After the embedding stage,
Each text 
is represented univocally by a sequence of vectors $\textEmbed = (\wordEmbed_1, \ldots, \wordEmbed_\textLength)$, where $m\in \mathbb{N}$,  $ \wordEmbed_i \in \textEmbedSet$, padding if necessary.
In this work we consider text classification with neural networks, hence, a text embedding $\textEmbed$ is classified into a category $\class \in \classes$, through a trained network $\network : \realNumber^{\embedDimen \cdot \textLength}_{[0,1]} \to \realNumber^{\abs{\classes}},$ i.e., $\class = \arg \max_{i \in \classes}\network_i(\textEmbed)$, where without any loss of generality we assume that each dimension of the input space of $\network$ is normalized between $0$ and $1$. We note that pre-trained embeddings are scaled before training, thus resulting in a L$_{\infty}$ diameter whose maximum value is 1. Thus, the lower and upper bound measurements are affected by normalization only when one compares embeddings with different dimensions with norms different from L$_{\infty}$.
In this paper robustness is measured for both convolutional and recurrent neural networks with the distance between words in the embedding space that is calculated with either L$_2$ or L$_{\infty}$-norm: while the former is a proxy for semantic similarity between words in polarized embeddings (this is discussed more in details in the Experimental Section), the latter, by taking into account the maximum variation along all the embedding dimensions, is used to compare different robustness profiles.

\subsection{Robustness Measure against Word Substitutions}
\begin{figure}[t]
    \centering
	\includegraphics[width=\linewidth,keepaspectratio]{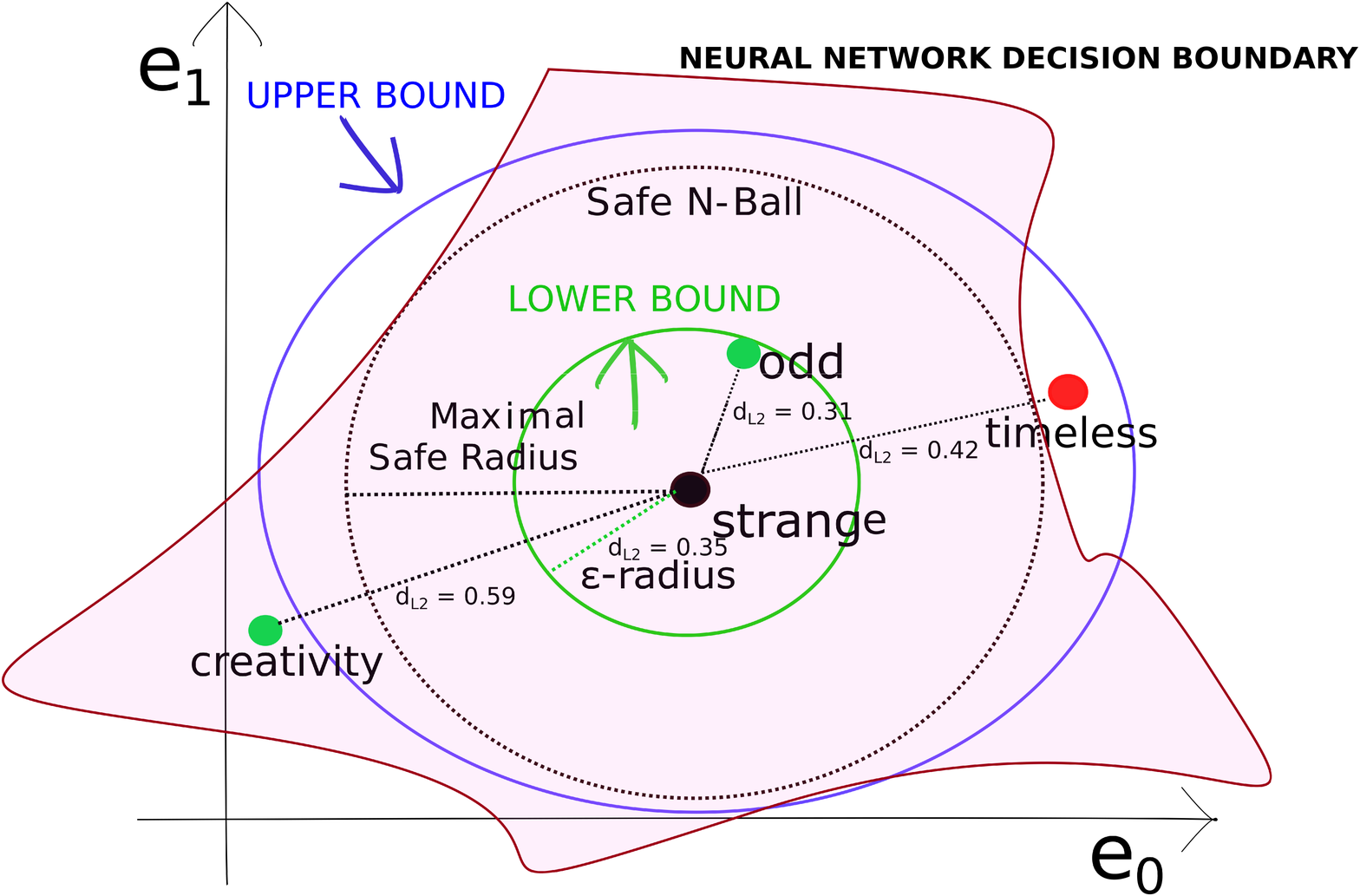}
	\caption{%\MK{We don't define MSR for words, only for text, so this is text with strange replaced by odd, etc}
	Illustration of the %characterization of the
	Maximal Safe Radius (MSR) and its upper and lower bounds. An upper bound of MSR is obtained by computing the distance of any perturbation resulting in a class change (blue ellipse) to the input text. A lower bound certifies that perturbations of the words contained within that radius are guaranteed to not change the classification decision (green ellipse). Both upper and lower bounds approximate the MSR (black ellipse). In this example the word $\texttt{strange}$ can be safely substituted with $\texttt{odd}$. The word $\texttt{timeless}$ is within upper and lower bound of the MSR, so our approach cannot guarantee it would not change the neural network prediction.% \MW{I changed the font of the words. Need to be consistent throughout the paper.}
	} 
	\label{fig:msr}
\end{figure}

Given a text embedding $\textEmbed$, a metric $L_p$, a subset of word indices $I\subseteq \{1,\ldots, m \}$, and a distance $\dist \in \realNumber_{\geq 0}$, we define $\ball(\textEmbed,  \dist) = \{ \textEmbed' \in \realNumber^{\embedDimen \cdot \textLength}_{[0,1]} \mid \norm{\textEmbed_I - \textEmbed_I'}_{p} \leq \dist \wedge (\forall  %\text{ for } 
i\notin I, \wordEmbed_i=\wordEmbed_i' ) \}$, where $\textEmbed_I$ is the sub-vector of $\textEmbed$ that contains only embedding vectors corresponding to words in $I$. That is, $\ball(\textEmbed,  \dist)$ is   the set of embedded texts obtained by replacing %perturbing
words in $I$ within $\textEmbed$ and whose distance to $\textEmbed$ is no greater than $\dist$. 
%\MK{We need the following or have to add the index set}
We elide the index set $I$ to simplify the notation. %, as it is clear from the context.
Below we define the notion of the \emph{maximal safe radius} (MSR), which is the minimum distance of an embedding text from the decision boundary of the network.
\begin{definition}[Maximal Safe Radius]
\label{def:MaximumSafeRadius}
    Given a neural network $\network$, a subset of word indices $I\subseteq \{1,\ldots,m \},$ and a text embedding $\textEmbed$, the \emph{maximal safe radius} $\MSR(\network, \textEmbed)$ is the minimum distance from input $\textEmbed$ to the decision boundary, i.e.,  $\MSR(\network, \textEmbed)$ is %\MK{I think we have to say MSR is the supremum or the largest epsilon, but this is less precise}\LL{Being the set of $\epsilon$ compact they should be the same.} 
    equal to the largest $\epsilon \in \mathbb{R}_{\geq 0} $ such that $\forall \textEmbed' \in \ball(\textEmbed, \dist) :\arg\max_{i}\network_{i\in C}(\textEmbed') = \arg\max_{i}\network_{i\in C}(\textEmbed) $.
\end{definition}
For a text $\textEmbed$  let $\mathbf{d}=\max_{\textEmbed' \in \realNumber^{\embedDimen \cdot \textLength}_{[0,1]} } \norm{\textEmbed_I - \textEmbed_I'}_{p}$ be the diameter of the embedding, then a large value for the normalised $\MSR$, $\frac{\MSR(\network, \textEmbed)}{\mathbf{d}}$, indicates that $\textEmbed$ is \emph{robust} to perturbations of the given subset $I$ of its words, as substitutions of these words do not result in a class change in the NN prediction (in particular, if the normalised $\MSR$ is greater than $1$ then $\textEmbed$ is robust to any perturbation of the words %with indices
in $I$). 
Conversely, low values of the normalised $\MSR$ indicate that the network's decision is  \emph{vulnerable at} $\textEmbed$ because of the ease with which the classification outcomes can be manipulated.
Further, averaging $\MSR$ over a set of inputs yields a \emph{robustness measure of the network}, as opposed to being specific to a given text. Under standard assumptions of %\MK{what is bounded variation?}\LL{replace problem with function, now it should be clearer} 
bounded variation of the underlying learning function, the $\MSR$ is also generally employed to quantify the robustness of the NN to adversarial examples~\cite{wugame:2019,weng2018towards}, that is, small perturbations that yield a prediction that differs from ground truth. Since computing the $\MSR$ is NP-hard~\cite{katzreluplex:2017}, we instead approximate it by computing a \emph{lower} and an \emph{upper} bound for this quantity (see Figure~\ref{fig:msr}). The strategy for obtaining an upper bound is detailed in Section \ref{sec:mcts}, whereas for the lower bound (Section \ref{Sec:lower bound}) we adapt constraint relaxation techniques developed for the verification of deep neural networks.

\subsection{Upper Bound: Monte Carlo Tree Search}
\label{sec:mcts}
\begin{figure}[t]
	\includegraphics[width=\linewidth, keepaspectratio]{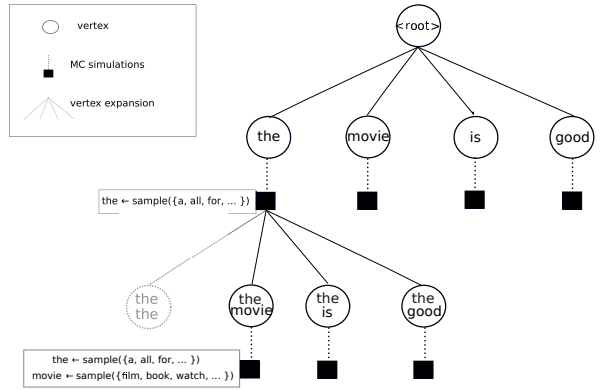}
	\caption{Structure of the tree after two iterations of the MCTS algorithm. Simulations of $1$-word substitutions are executed at each vertex on the first level to update the UCT statistics. The most urgent vertex is then expanded (e.g., word $\texttt{the}$) and several $2$-words substitutions are executed combining the word identified by the current vertex (e.g., word $\texttt{movie}$ at the second level of the tree) and that of its parent, i.e.,  $\texttt{the}$. Redundant substitutions may be avoided (greyed out branch).
	} 
	\label{fig:mcts}
\end{figure}

An upper bound for $\MSR$ is a perturbation of the text that is classified by the NN differently than the original text. In order to only consider perturbations that are syntactically coherent with the input text, we use filtering in conjunction with an adaptation of the Monte Carlo Tree Search (MCTS) algorithm~\cite{coulommcts:2007} to the NLP scenario  (Figure~\ref{fig:mcts}).
The algorithm takes as input a \emph{text}, embeds it as a sequence %list 
of vectors $\textEmbed$, and builds a tree where at each iteration a set of indices $I$ identifies the words that have been modified so far: at the first level of the tree a single word is changed to manipulate the classification outcome, at the second two  words are perturbed, with the former being the same word as for the parent vertex, and so on (i.e., for each vertex, $I$ contains the indices of the words that have been perturbed plus that of the current vertex). We allow only word for word substitutions. 
At each stage the procedure outputs all the successful attacks (i.e., perturbed texts that are classified by the neural network differently from the original text) that have been found until the terminating condition is satisfied (e.g., a fixed fraction out of the total number of vertices has been explored). Successful perturbations can be used as diagnostic information in cases where ground truth information is available. The algorithm explores the tree according to the UCT heuristic~\citep{browne2012survey}, where \textit{urgent} vertices are identified by the perturbations that induce the largest drop in the neural network's confidence.
A detailed description of the resulting algorithm, which follows the classical
algorithm~\cite{coulommcts:2007} while working directly with word embeddings, can be found in Appendix~\ref{app:mcts}.
Perturbations are sampled by  considering the $n$-closest replacements in the word's neighbourhood: the distance between words is measured in the L$_2$ norm, while the number of substitutions per word is limited to a fixed constant (e.g., in our experiments this is either $1000$ or $10000$). In order to enforce the syntactic consistency of the replacements we consider part-of-speech tagging of each word %as a noun, verb, adjective, etc., 
based on its context. Then, we filter all the replacements found by MCTS to exclude those that are not of the same type, or from a type that will maintain the syntactic consistency of the perturbed text (e.g., a noun sometimes can be replaced by an adjective). % without making the phrase wrong).
To accomplish this task we use the Natural Language Toolkit~\citep{bird2009natural}.
More details are provided in Appendix~\ref{app:subs}.

\subsection{Lower Bound: Constraint Relaxation}
\label{Sec:lower bound}
A lower bound for $\MSR(\network, \textEmbed)$ is a real number $\epsilon_l>0$ such that all texts in $\ball(\textEmbed,  \epsilon_l)$  are classified in the same class by $\network$.  Note that, as $\MSR(\network, \textEmbed)$ is defined in the embedding space, which is \emph{continuous}, the perturbation space, $\ball(\textEmbed,\epsilon)$, contains meaningful texts as well as texts that are not syntactically or semantically meaningful.
In order to compute $\epsilon_l$ we %\MK{should be 'we leverage constraint relaxation', not 'leverage on'}\LL{My fault :)}  
leverage  constraint relaxation techniques developed for CNNs~\cite{boopathy2019cnn} and LSTMs~\cite{kopopqorn:2019}, namely CNN-Cert and POPQORN. For an input text $\textEmbed$ and a hyper-box around $\ball(\textEmbed,\dist)$,  these techniques find linear lower and upper bounds for the activation functions of each layer of the neural network and use these to propagate an over-approximation of the hyper-box through the network. $\epsilon_l$ is then computed as the largest real such that all the texts in $\ball(\textEmbed,\epsilon_l)$ are %propagated 
in the same class, i.e., for all $\textEmbed'\in \ball(\textEmbed,\epsilon_l)$, $ \arg \max_{i \in \classes}\network_i(\textEmbed)= \arg \max_{i \in \classes}\network_i(\textEmbed')$. Note that,  as $\ball(\textEmbed,\epsilon_l)$ contains only texts obtained by perturbing a subset of the words (those whose index is in $I$), to adapt CNN-Cert and POPQORN to our setting, we have to fix the dimensions of $\textEmbed$ corresponding to words not in $I$ and only propagate through the network intervals  corresponding to words in $I.$

\section{Experimental Results}
\label{sec:experiments}
\begin{table*}[t]
\centering
\vspace{2 mm}
\scalebox{0.9}{
\begin{tabular}{|c|c|c|c|c|}
\hline
 \textbf{} &  \small{\textbf{NEWS}} & \small{\textbf{SST}} &  \small{\textbf{AG NEWS}} & \small{\textbf{IMDB}} \\ \hline
  \multirow{1}{*}{\small{\textbf{Inputs (Train, Test)}}} & \small{$22806,9793$} &  \small{$117220,1821$} & \small{$120000,7000$} & \small{$25000,25000$} \\ \cline{1-5}
  \multirow{1}{*}{\small{\textbf{Output Classes}}} & \small{$7$} &  \small{$2$} & \small{$4$} & \small{$2$} \\ \cline{1-5}
   \multirow{1}{*}{\small{\textbf{Average Input Length}}} & \small{$17 \pm 2.17$} &  \small{$17.058 \pm 8.27$} & \small{$37.295 \pm 9.943$} & \small{$230.8 \pm 169.16$} \\ \cline{1-5}
  \multirow{1}{*}{\small{\textbf{Max Input Length}}} & \small{$88$} &  \small{$52$} & \small{$136$} & \small{$2315$} \\ \cline{1-5}   
  \multirow{1}{*}{\small{\textbf{Max Length Considered}}} & \small{$14$} &  \small{$25$} & \small{$49$} & \small{$100$} \\ \cline{1-5}  
\end{tabular} }
\caption{Datasets used for the experimental evaluation.
We report the number of samples (training/test ratio as provided in the original works) and output classes, the average and maximum length of each input text before pre-processing and the maximum length considered in our experiments.
}
\label{tab:data}
\end{table*}
We use our framework to empirically evaluate the robustness of neural networks for sentiment analysis and news classification on typical CNN and LSTM architectures. While we quantify lower bounds of $\MSR$ for CNNs and LSTMs, respectively, with CNN-Cert and POPQORN tools, we implement the MCTS algorithm introduced in Section~\ref{sec:mcts} to search for meaningful perturbations (i.e., upper bounds), regardless of the NN architecture employed.
In particular, in Section \ref{subsec:words_substitution} we consider robustness against single and multiple word substitutions and investigate implicit biases of LSTM architectures. In Section~\ref{subsec:embeddings_and_robustness} we study the effect of embedding on robustness, while in Section \ref{subsec:interpretability} we employ our framework to perform saliency analysis of the most relevant words in a text.
\paragraph{Experimental Setup and Implementation} \label{sec:experiments-setup}
We have trained several vanilla CNN and LSTM models on datasets that differ in length of each input, number of target classes and difficulty of the learning task. All our experiments were conducted on a server equipped with two $24$ core Intel Xenon $6252$ processors and $256$GB of RAM\footnote{We emphasise that, although the experiments reported here have been performed on a cluster, all the algorithms are reproducible on a mid-end laptop; we used a machine with 16GB of RAM and an Intel-5 8th-gen. processor.}\textsuperscript{,}\footnote{Code for reproducing the MCTS experiments is available at: \url{https://github.com/EmanueleLM/MCTS}}. We consider the IMDB dataset~\citep{maaslearning:2011}, the Stanford Sentiment Treebank (SST) dataset~\citep{socherrecursive:2013}, the AG News Corpus~\citep{zhangcharacter:2015} and the NEWS dataset~\citep{vitale2012classification}: details are in Table~\ref{tab:data}.
In our experiments we consider different embeddings, and specifically both complex, probabilistically-constrained representations (GloVe and GloVeTwitter) trained on global word-word co-occurrence statistics from a corpus, as well as 
the simplified embedding provided by the Keras Python Deep Learning Library (referred to as Keras Custom)~\citep{chollet2015keras}, 
which allows one to fine tune the exact dimension of the vector space and only aims at minimizing the loss on the classification task. The resulting learned Keras Custom embedding does not capture complete word semantics, just their emotional polarity. More details are reported in Appendix~\ref{app:other-results} and Table~\ref{tab:embeddings}. For our experiments, we consider a $3$-layer CNN,  where the first layer consists of bi-dimensional convolution with $150$ filters, each of size $3 \times 3$, and a LSTM model with $256$ hidden neurons on each gate. %Considering the embeddings,
We have trained more than $20$ architectures on the embeddings and datasets mentioned above.  %introduced previously. 
We note that, though other architectures might offer higher accuracy for sentence classification~\citep{kim-2014-convolutional}, this \textit{vanilla} setup has been chosen intentionally not to be optimized for a specific task, thus allowing us to measure robustness of baseline models. %; with the same rationale we have chosen a standard LSTM model with $256$ hidden neurons on each gate. 
Both CNNs and LSTMs predict the output with a $\mathsf{softmax}$ output layer, while the $\mathsf{categorical}$ $\mathsf{cross}$-$\mathsf{entropy}$ loss function is used during the optimization phase, which is performed with Adam~\cite{kingma2014adam} algorithm (without early-stopping); further details are reported in Appendix~\ref{app:other-results}.

\subsection{Robustness to Word Substitutions} \label{subsec:words_substitution}
\begin{figure*}[t]
    \centering
	\includegraphics[width=\linewidth, keepaspectratio]{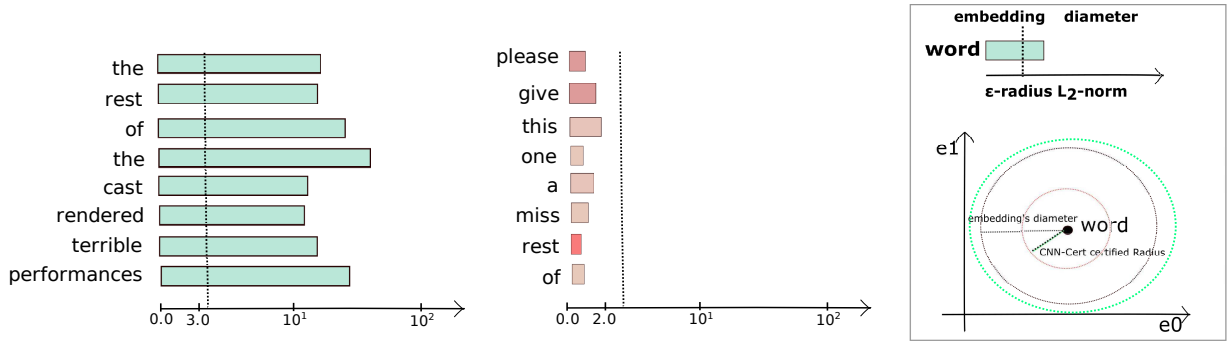}
	\caption{ Lower bounds indicate classification invariance to any substitution when greater than the embedding diameter  $\mathbf{d}$ (see diagram on the right and Section \ref{sec:text-classification}), here represented by the dotted vertical line. Left: Examples of words safe to any substitution (IMDB, Keras embedding $10d$, text no 2). Middle: Examples of words vulnerable to substitutions that may change the classification (IMDB, Keras embedding $5d$, text no 1). 
	} 
	\label{fig:safety}
\end{figure*}

\begin{figure}[t]
    \centering
	\includegraphics[width=\linewidth, keepaspectratio]{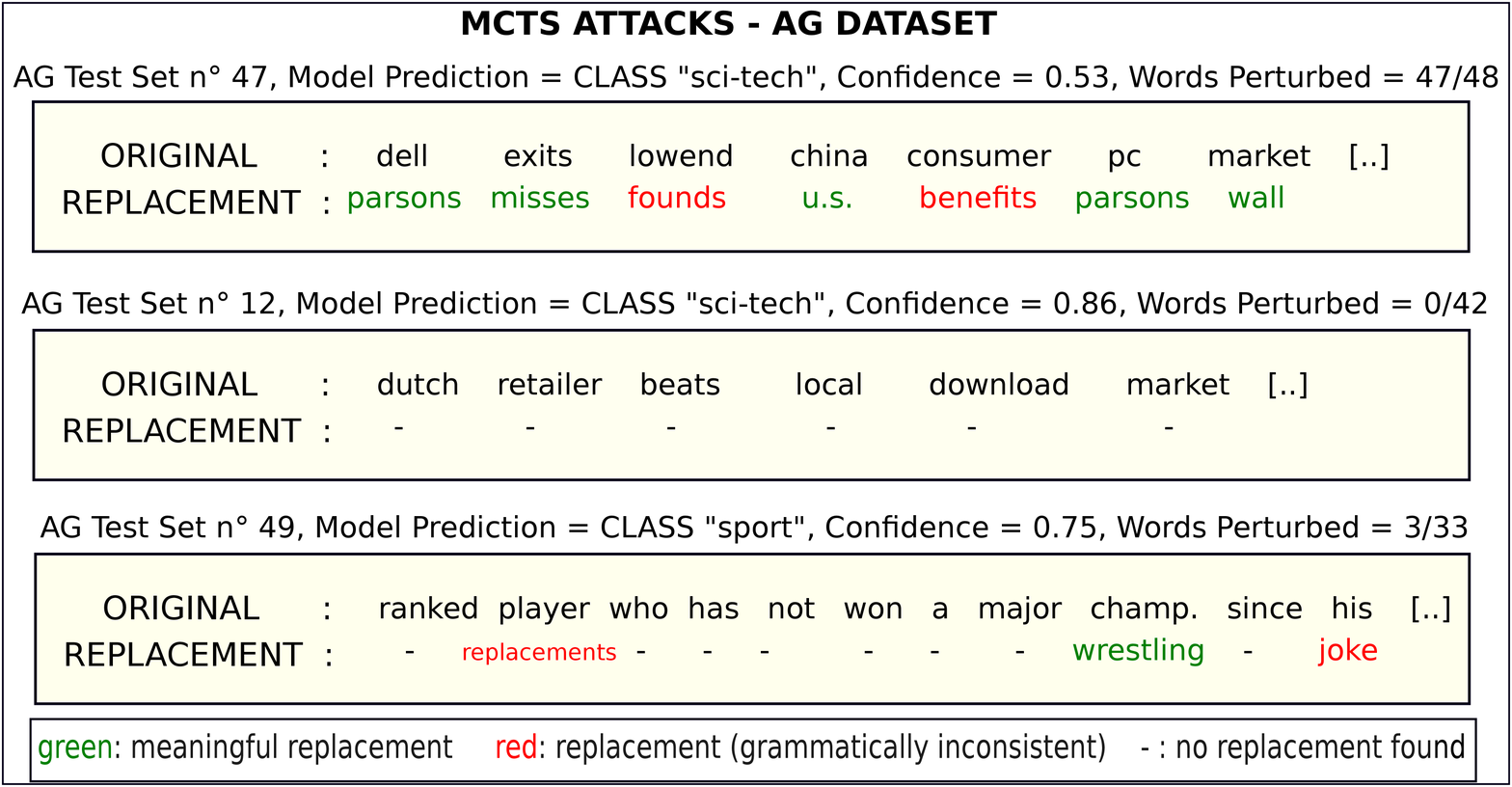}
	\caption{Single-word substitutions found with MCTS in conjunction with filtering. Grammatically consistent substitutions shown in green, inconsistent in red, a dash indicates that no substitution is found.} 
	\label{fig:mcts_meaningful_subs}
\end{figure}

For each combination of a neural network and embedding, we quantify the $\MSR$ against single and multiple word substitutions, meaning that the set of word indices $I$ (see Definition \ref{def:MaximumSafeRadius}) consists of $1$ or more indices.
Interestingly, our framework is able to prove that certain input texts and architectures are robust  for any single-word substitution,
that is, replacing a single word of the text (any word) with any other possible other word, and not necessarily with a synonym or a grammatically correct word, will not affect the classification outcome. %An example is shown in
Figure~\ref{fig:safety} shows that for CNN models {equipped} with Keras Custom embedding the (lower bound of the) $\MSR$  on some %test sets 
texts 
from the IMDB dataset is greater than the diameter of the embedding space. 
\begin{table}[t]
\centering
{}
\vspace{2 mm}
\scalebox{0.8}{
\begin{tabular}{|c|c|c|c|}
\hline
 \textbf{} &  \small{\textbf{DIMENSION}} &  \small{\textbf{LOWER BOUND}} \\ \hline
  \multirow{6}{*}{\textbf{Keras}} & \small{$5$} &  \small{$0.278$}  \\ \cline{3-3}
                          & \small{$10$} & \small{$0.141$} \\ \cline{3-3}
                          & \small{$25$} & \small{$0.023$} \\ \cline{3-3}
                          & \small{$50$} & \small{$0.004$} \\ \cline{3-3}
                          & \small{$100$} & \small{$0.002$} \\ 
                          \hline
  \multirow{2}{*}{\textbf{GloVe}} & \small{$50$} & \small{$0.007$} \\ \cline{3-3}
                         & \small{$100$} & \small{$0.002$} \\ \hline
  \multirow{3}{*}{\textbf{GloVeTwitter}} & \small{$25$} & \small{$0.013$} \\ \cline{3-3}
                                & \small{$50$} & \small{$0.008$} \\ \cline{3-3}
                                & \small{$100$} & \small{$0.0$}\\ \hline
\end{tabular} }
\\
\caption{Comparison of lower bounds for single-word substitutions computed by CNN-Cert on the SST dataset. Values are averaged over $100$ input texts (approx. $2500$ measurements) and normalized by the embedding diameter (L$_2$-norm).
}
\label{tab:sst_table}
\end{table}
To consider only perturbations that are semantically close and
syntactically coherent with the input text,  we employ the  MCTS algorithm with filtering described in Section~\ref{sec:mcts}. 
An example of a successful perturbation is shown in Figure~\ref{fig:mcts_meaningful_subs}, where we illustrate the effectiveness of single-word substitutions on inputs that differ in the confidence of the neural network prediction. We note that even with simple tagging it is possible to identify perturbations where replacements are meaningful. For the first example in Figure~\ref{fig:mcts_meaningful_subs} (top), the network changes the output class to \texttt{World} when the word \texttt{China} is substituted for \texttt{U.S.}. Although this substitution may be relevant to that particular class, nonetheless we note that the perturbed text is coherent and the main topic remains \texttt{sci}-\texttt{tech}. Furthermore, the classification changes also when the word \texttt{exists} is replaced with a plausible alternative \texttt{misses}, a perturbation that is %semantically coherent and apparently 
\textit{neutral}, i.e. %this term is 
not informative for any of the possible output classes. In the third sentence in Figure~\ref{fig:mcts_meaningful_subs} (bottom), we note that replacing \texttt{championship} with \texttt{wrestling} makes the model output class \texttt{World}, where originally it was \texttt{Sport}, indicating that the model relies on a small number of key words to make its decision. % : differently from the first input, here we note that few words allow for effective perturbations, a hint that on this sentence the model relies on few key-words to make its decision.
We report a few additional examples of word replacements for a CNN model equipped with GloVe-50d embedding. Given as input the review \texttt{’this is art paying homage to art'} (from the SST dataset), when \texttt{art} is replaced by \texttt{graffiti} the network misclassifies the review (from \textit{positive} to \textit{negative}). Further, as mentioned earlier, the MCTS framework is capable of finding multiple word perturbations: considering the same setting as in the previous example, when in the review \texttt{'it’s not horrible just horribly mediocre'} the words \texttt{horrible} and \texttt{horribly} are replaced, respectively, with \texttt{gratifying} and \texttt{decently}, the review is classified as \textit{positive}, while for the original sentence it was \textit{negative}. 
Robustness results for high-dimensional embeddings are included in Table~\ref{tab:msr_gap_table}, where we report the trends of the  average lower and upper bounds of $\MSR$ and the percentage of successful perturbations computed over $100$ %different input
texts (per dataset) for %the various
different architectures and embeddings.
Further results are in Appendix~\ref{app:other-results}, including statistics on lower bounds (Tables~\ref{tab:imdb_sst_news_table},~\ref{tab:ag_table}) and %the MCTS algorithm (
single and multiple word substitutions %, respectively in
(Tables~\ref{tab:mcts_table},~\ref{tab:mcts_table_multiple_attacks}).

\begin{table*}[t]
\centering
{Single-Word Substitutions}
\\
\vspace{2 mm}
\scalebox{0.9}{
\begin{tabular}{|c|c|c|c|c|c|}
\hline
& {\textbf{\small{EMBEDDING}}} & {\textbf{\small{LOWER BOUND}}} & \multicolumn{2}{c|}{\textbf{\small{SUBSTITUTIONS}}} & {\textbf{\small{UPPER BOUND}}} \\ \cline{4-5}
  & {} & {} & \textbf{\small{\% per text}} & \textbf{\small{\% per word}} & {} \\ \hline
  \multirow{3}{*}{\textbf{\small{IMDB}}} & \small{Keras$50$d} & \small{$0.055\pm0.011$}  &  \small{$6.0$}  & \small{$1.4$} & \small{$0.986$}  \\ \cline{3-6}
  & \small{GloVe$50$d} & \small{$0.018\pm0.007$}  &  \small{$39.7$}  & \small{$5.1$} & \small{$0.951$}  \\ \cline{3-6}
  & \small{GloVeTwitter$50$d} & \small{$0.02\pm0.002$}  &  \small{$47.0$}  & \small{$7.7$} & \small{$0.926$} \\ \hline  
  \multirow{3}{*}{\textbf{\small{AG News}}} & \small{Keras$50$d} & \small{$0.002\pm0.001$} &  \small{$50.0$}  & \small{$15.6$}  & \small{$0.852$}  \\ \cline{3-6}
  & \small{GloVe$50$d} & \small{$0.005\pm0.004$}  &  \small{$22.4$}  & \small{$10.8$} & \small{$0.898$} \\ \cline{3-6}
  & \small{GloVeTwitter$50$d} & \small{$0.007\pm0.001$}  &  \small{$21.4$}  & \small{$6.6$} & \small{$0.937$}  \\ \hline
  \multirow{3}{*}{\textbf{\small{SST}}} & \small{Keras$50$d} & \small{$0.004\pm0.001$}  &  \small{$52.2$}  & \small{$19.9$} & \small{$0.813$} \\ \cline{3-6}
  & \small{GloVe$50$d} & \small{$0.007\pm0.003$}  &  \small{$81.1$}  & \small{$37.4$} & \small{$0.646$} \\ \cline{3-6}
  & \small{GloVeTwitter$50$d} & \small{$0.008\pm0.004$}  &  \small{$78.1$}  & \small{$36.3$} & \small{$0.653$}  \\   \hline 
  \multirow{4}{*}{\textbf{\small{NEWS}}} & \small{GloVe$50$d} & \small{$0.001\pm0.002$} &  \small{$96.5$}  & \small{$34.0$}  & \small{$0.679$} \\ \cline{3-6}
  & \small{GloVe$100$d} & \small{$0.002 \pm 0.002$}  &  \small{$89.7$}  & \small{$29.1$} & \small{$0.727$} \\ \cline{3-6}
  & \small{GloVeTwitter$50$d} & \small{$0.001\pm0.001$}  &  \small{$90.9$}  & \small{$30.6$} & \small{$0.707$}   \\ \cline{3-6}  
  & \small{GloVeTwitter$100$d} & \small{$0.001\pm0.001$}  &  \small{$89.7$}  & \small{$27.7$} & \small{$0.739$}  \\  \hline                    
\end{tabular} }
\\
\caption{
Statistics on single-word substitutions averaged on $100$ input texts of each dataset. 
We report: the average lower bound of the $\MSR$ as measured with either CNN-Cert or POPQORN; the  approximate ratio that given a word from a text we find a single-word substitution and the average number of words that substituted for a given word change the classification; the average upper bound computed as the distance between the original word and the closest substitution found by MCTS (when no successful perturbation is found we over-approximate the upper bound for that word with the diameter of the embedding). Values reported for lower bounds have been normalized by each embedding diameter (measurements in the L$_2$-norm).
}
\label{tab:msr_gap_table}
\end{table*}

\begin{figure}[t]
    \centering
	\includegraphics[width=\linewidth, keepaspectratio]{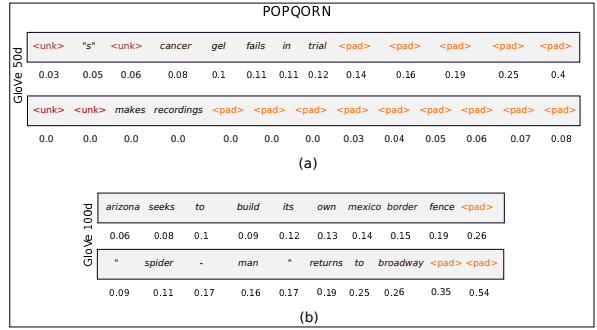}
	\caption{Lower bound values for individual words obtained from POPQORN ($L_2$-norm), showing an increasing trend for consecutive words. (a) Two texts with padding (\texttt{<unk>} denotes an unknown token). (b) Texts with several words related to a specific output class (\texttt{U.S.} and \texttt{entertainment}, respectively).
	} 
	\label{fig:popqorn_polarization}
\end{figure}

\begin{figure}[t]
    \centering
	\includegraphics[width=\linewidth, keepaspectratio]{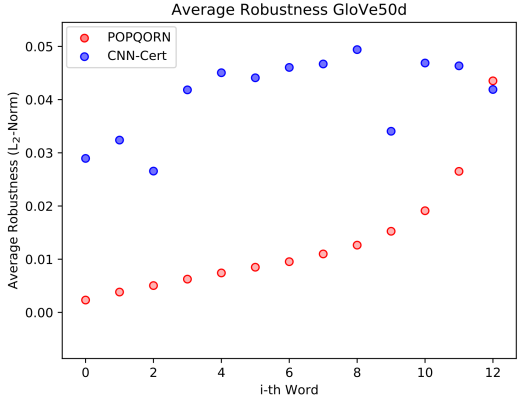}
	\caption{Robustness lower bound trends for successive input words for LSTMs (red dots) and CNNs (blue dots) on NEWS and AG News datasets.
	} 
	\label{fig:avg_robustness_cnncert_vs_popqorn}
\end{figure}

\paragraph{CNNs vs. LTSMs} \label{subsec:implicit_biases}
By comparing the average robustness assigned to
 each word, respectively, by CNN-Cert and POPQORN over all the experiments on a fixed dataset, it clearly emerges that recurrent models are less robust to perturbations that occur in very first words of a sentence; interestingly, CNNs do not suffer from this problem. A visual comparison is shown in Figure~\ref{fig:avg_robustness_cnncert_vs_popqorn}. The key difference is the structure of LSTMs compared to CNNs: while in LSTMs the first input word  influences the successive layers, thus amplifying the manipulations,
the output of a convolutional region is independent from any other of the same layer.
On the other hand, both CNNs and LSTMs have in common an increased resilience to perturbations on texts that contain multiple polarized words, a trend that suggests that, independently of the architecture employed, robustness relies on a distributed representation of the content in a text (Figure~\ref{fig:popqorn_polarization}).

\subsection{Influence of the Embedding on Robustness} \label{subsec:embeddings_and_robustness}
As illustrated in Table~\ref{tab:sst_table} and in Figure~\ref{fig:safety}, models that employ small embeddings are more robust to perturbations. On the contrary, robustness decreases, from one to two orders of magnitude, when words are mapped to high-dimensional spaces, a trend that is confirmed also by MCTS  (see Appendix Table~\ref{tab:mcts_table_multiple_attacks}). This may be explained by the fact that adversarial perturbations are inherently related to the dimensionality of the input space~\citep{carbone2020robustness,goodfellowexplaining:2014}.
We also discover that models trained on longer inputs (e.g., IMDB) are more robust compared to those trained on shorter ones (e.g., SST): in long texts the decision made by the algorithm depends on multiple words that are evenly distributed across the input, while for shorter sequences the decision may depend on very few, polarized terms.
From Table~\ref{tab:msr_gap_table} we note that polarity-constrained embeddings (Keras) are more robust than those that are probabilistically-constrained (GloVe) on relatively large datasets (IMDB), whereas the opposite is true on smaller input dimensions: experiments suggest that models with embeddings that group together words closely related to a specific output class (e.g., \textit{positive} words) are more robust, as opposed to models whose embeddings gather words together on a different principle (e.g., words that appear in the same context): intuitively, in the former case, words like \texttt{good} will be close to synonyms like \texttt{better} and \texttt{nice}, while in the latter words like \texttt{good} and \texttt{bad}, which often appear in the same context (think of the phrase \texttt{'the movie was good/bad'}), will be closer in the embedding space.
In the spirit of the analysis in %Baroni et al.
\citep{baroni-etal-2014-dont}, we empirically measured whether robustness is affected by the \textit{nature} of the embedding employed, that is, either prediction-based (i.e., embeddings that are trained alongside the classification task) or hybrid/count-based (e.g., GloVe, GloVeTwitter). By comparing the robustness of different embeddings and the distance between words that share the same polarity profile (e.g., \textit{positive} vs. \textit{negative}), we note that $\MSR$ is  a particularly well suited robustness metric for prediction-based embeddings, with the distance between words serving as a reasonable estimator of word-to-word semantic similarity w.r.t. the classification task. On the other hand, for hybrid and count-based embeddings (e.g., GloVe), especially when words are represented as high-dimensional vectors, the distance between two words in the embedding space, when compressed into a single scalar, does not retain enough information to estimate the relevance of input variations. Therefore, in this scenario,  an approach based solely on the $\MSR$ is limited by the choice of the distance function between words, and may lose its effectiveness unless additional factors such as context are considered.
Further details of our evaluation are provided in Appendix~\ref{app:other-results}, Table~\ref{tab:imdb_sst_news_table} and Figure~\ref{fig:epsreg}.

\paragraph{Counter-fitting} \label{subsec:counter-fitting}
 \begin{figure}[t]
    \centering
	\includegraphics[width=\linewidth,keepaspectratio]{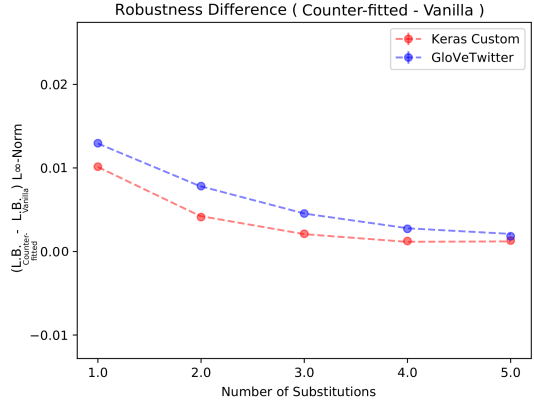}
	\caption{For an increasing number of substitutions per text we report the difference between  $\MSR$ lower bounds  %as computed by our framework 
	of counter-fitted and %counter vs. non 
	vanilla embeddings (Keras and GloVeTwitter, $25$d) on the AG News Dataset. 
	}
	\label{fig:cfittting_multiple_attacks}
\end{figure}
To mitigate the issue of robustness in multi-class datasets characterized by short sequences, we have repeated the robustness measurements with counter-fitted~\citep{mrkvsiccounter:2016} embeddings, i.e., a method of injecting additional constraints for antonyms and synonyms   %\textit{"antonymy and synonymy constraints [are injected] 
into vector space representations in order  to  improve  the  vectors’  capability  to encode  semantic  similarity. %\MK{Describe what counter-fitting is in one sentence}\ELM{Done}
We observe that the estimated lower bound of $\MSR$ is in general increased for low-dimensional embeddings, up to twice the lower bound for non counter-fitted embeddings. This phenomenon is particularly relevant when Keras Custom $5$d and $10$d are employed, see Appendix~\ref{app:other-results}, Table~\ref{tab:ag_table}. On the other hand, the benefits of counter-fitting are less pronounced for high-dimensional embeddings.  %On the contrary, we note that counter-fitting does not give noticeable help when the embedding \MK{dimensionality not size?} size grows larger, as the benefits induced are balanced by the high dimensionality of the input space. 
The same pattern can be observed in Figure~\ref{fig:cfittting_multiple_attacks}, where   multiple-word substitutions per text are allowed. %\MK{I don't understand, also Fig 7}\LL{I tried to modify the text. We should specify the dimension of the embedding in the caption}. 
Further details can be found in Appendix~\ref{app:other-results}, Tables~\ref{tab:ag_table} and~\ref{tab:mcts_table_multiple_attacks}.

\subsection{Interpretability of Sentiment Analysis via Saliency Maps} \label{subsec:interpretability}
\begin{figure}[t]
    \centering
	\includegraphics[width=\linewidth, keepaspectratio]{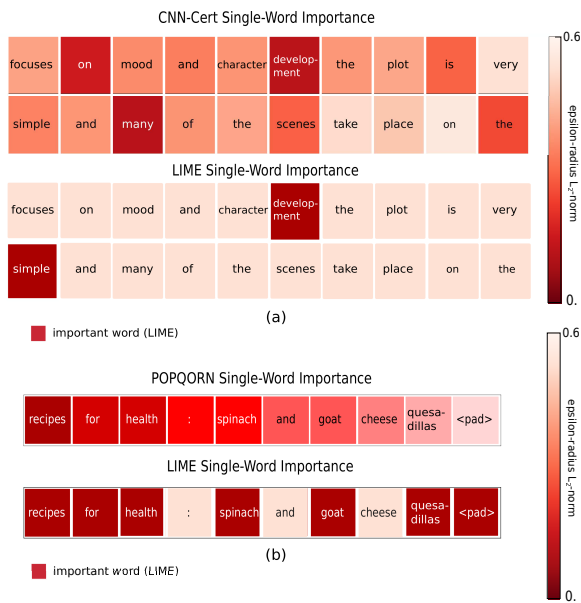}
	\caption{Interpretability comparison of our framework with LIME. 
	(a) Saliency map produced with CNN-Cert (top) and LIME (bottom) on IMDB (GloVeTwitter $25$d embedding).
	(b) Saliency map produced with POPQORN (top) and LIME (bottom) on NEWS dataset (GloVe $100$d embedding).
% 	See Section \ref{subsec:interpretability} for a discussion.
	} 
	\label{fig:lime}
\end{figure}
We employ our framework to perform interpretablity analysis on a given text. %, that is, to quantify the importance of each word in the text. In particular, 
For each word of a given text we compute the (lower bound of the) $\MSR$ and use this as a measure of its \emph{saliency}, %how much that word is important for the classification problem, i.e.,
where small values of $\MSR$ indicate that minor perturbations of that word can have a significant influence on the classification outcome.
We use the above measure to compute saliency maps for both CNNs and LSTMs, and compare our results with those obtained  by  LIME~\citep{lime:2016}, which assigns saliency %importance 
to input features according to the best linear model that locally  explains the decision boundary.
Our method has the advantage of being able to account for non-linearities in the decision boundary that a local approach such as LIME cannot handle, albeit at a cost of higher computational complexity (a similar point was made in \cite{blaas2019robustness} for Gaussian processes). 
As a result, we are able to discover words that our framework views as important, but LIME does not, and vice versa.
%\MK{Make names of datasets consistent, eg NEWS vs AG News}%\ELM{Now it is consistent}
In Figure \ref{fig:lime} we report two examples, one for an IMDB \textit{positive} review (Figure~\ref{fig:lime} (a)) and another from the NEWS dataset classified using a LTSM (Figure~\ref{fig:lime} (b)).
In Figure~\ref{fig:lime} (a) our approach %based on computing the $\MSR$ via CNN-cert we 
finds that the word \texttt{many} % ($2^{nd}$ row)
is salient and perturbing it slightly can make the NN change the class of the review to \textit{negative}. In contrast, LIME does not identify \texttt{many} as significant. In order to verify this result empirically, we run our MCTS algorithm (Section \ref{sec:mcts}) and find that simply substituting \texttt{many} with \texttt{worst} changes the classification to %of the NN, i.e., with this only change the review is classified as a 
`negative'. Similarly, for Figure~\ref{fig:lime} (b), where the input is assigned to class $5$ (\texttt{health}), perturbing the punctuation mark (\texttt{:}) may alter the classification, whereas LIME does not recognise its saliency. 

\section{Conclusions}
We introduced a framework for evaluating robustness of NLP models against word substitutions.
Through extensive experimental evaluation we demonstrated that our framework allows one to certify certain architectures against single word perturbations and illustrated how it can be employed for interpretability analysis. While we focus on perturbations that are syntactically coherent, we acknowledge that semantic similarity between phrases is a crucial aspect that nonetheless requires an approach which takes into account the context where substitutions happen: we will tackle this limitation in future.  
Furthermore, we will address robustness of more complex architectures,
e.g., networks that exploit attention-based mechanisms~ \citep{vaswani2017attention}.

\subsubsection*{Acknowledgements}
This project was partly funded by Innovate UK (reference 104814) and the ERC under the European Union’s Horizon 2020 research and innovation programme (grant agreement No. 834115). 
We thank the reviewers for their critical assessment and suggestions for improvement.

\bibliography{anthology,emnlp2020}
\bibliographystyle{acl_natbib}

\newpage
\clearpage

\appendix
\section{Appendix}

\subsection{Monte Carlo Tree Search (MCTS)} \label{app:mcts}
%\MK{Need to cite the paper where the MCTS algorithm is taken from, then briefly say how it is adapted }%BB{Added citation to a paper that present a (smaller) version of the pesudo-code, plus significant difference with standard MCTS.}
We adapt the MCTS algorithm~\cite{browne2012survey} to the NLP classification setting with word embedding, %without introducing relevant variations to the standard algorithm: only slight adjustments are made to adapt each sub-routine to the NLP setting. In
which we report here for completeness as Algorithm~\ref{alg:mcts}.  %pseudo-code is reported. 
The algorithm explores modifications to the original text by substituting one word at the time with nearest neighbour alternatives. It takes as input: \textit{text}, expressed as a list of $T$ words; $\network$, the neural network as introduced in Section~\ref{sec:text-classification};  
$\embedding$, an embedding; \textit{sims}, an integer specifying the number of Monte Carlo samplings at each step; and $\alpha$, a real-valued meta-parameter specifying the exploration/exploitation trade-off for vertices that can be further expanded.
%\MK{But urgent is not mentioned in the description}%BB{True, the meaning of urgent is not specified: I thought it was a specific terminology as I found this notation (without any further explanation) on the paper "Evaluating the Complexity of Players' Strategies using MCTS Iterations", Lanzi P., IEEE}
The salient steps of the MCTS procedure are:
\begin{itemize}
    \item \textbf{Select}: the most \textit{promising} vertex to explore is chosen to be expanded (Line~\ref{lst:line:uctheuristic}) according to the standard UCT heuristic: \\ $\dfrac{Q(v)}{N(v)} + \alpha\sqrt{\dfrac{2lnN(v')}{N(v)}}$, where
    $v$ and $v'$ are respectively the selected vertex and its parent; $\alpha$ is a meta-parameter that balances exploration-exploitation trade-off;
    $N()$ represents the number of times a vertex has been visited; and $Q()$ measures the neural network confidence drop, averaged over the Monte Carlo simulations for that specific vertex.
    \item \textbf{Expand}: the tree is expanded with $T$ new vertices, one for each word in the input text (avoiding repetitions). A vertex at index $t \in \{1,... T\}$ and depth $n>0$ represents the strategy of perturbing the $t$-th input word, plus all the words whose indices have been stored in the parents of the vertex itself, up to the root.
    \item \textbf{Simulate}: simulations are run from the current position in the tree to estimate how the neural network behaves against the perturbations sampled at that stage (Line~\ref{lst:line:sample}). If one of the word substitutions induced by the simulation makes the network change the classification, a successful substitution is found and added to the results, while the value $Q$ of the current vertex is updated. Many heuristics can be considered at this stage, for example the average drop in the confidence of the network over all the simulations. We have found that the average drop is not a good measure of how the robustness of the network drops when some specific words are replaced, since for a high number of simulations a perturbation that is effective might pass unnoticed. We thus work with the maximum drop over all the simulations, which works slightly better in this scenario (Line~\ref{lst:line:simulate2}).
    \item \textbf{Backpropagate}: the  reward  received  is  back-propagated  to  the  vertices  visited  during  selection  and  expansion to update their UCT statistics. It is known that, when  UCT is employed~\citep{browne2012survey,kocsisbandit:2006}, MCTS guarantees that the probability of selecting a sub-optimal perturbation tends to zero at a polynomial rate when the number of games grows to infinity (i.e., it is guaranteed to find a discrete perturbation, if it exists).
\end{itemize}

\begin{algorithm*}
\caption{Monte Carlo Tree Search with UCT heuristic}\label{alg:mcts}
\begin{algorithmic}[1]

\Procedure{MCTS}{$text,\ \network,\ \embedding ,\ sims,\ \alpha$}
\State $t \gets \arg \max_{i \in \classes}\network_i(\embedding(text))$ \Comment{Store the unperturbed network output, ref. Section\ref{sec:text-classification}}
\State $Tree \gets createTree(text,\ c,\ \network)$  \Comment{Create the initial tree}
\State $root \gets getRoot(Tree)$  \Comment{Store the initial vertex}
\State $P \gets [ \ ]$ \Comment{List of final perturbations}
\While{$terminate(Tree)\not=True$}  \Comment{Loop over the MCTS steps}
\State $v \gets $SELECT$(Tree,\ \alpha)$
\State $C \gets $EXPAND$(v,\ text)$
\State $P.insert($SIMULATE$(C,\ text,\ sims,\ \network,\ \embedding,\ t))$
\State BACKPROPAGATE$(v, root)$
\EndWhile
\State \textbf{return} $P$
\EndProcedure

\item[]

\Procedure{Select}{$Tree,\ \alpha$}
\State $L \gets getLeaves(Tree)$
\State \textbf{return} $\operatorname*{argmax}_{v \in L} \dfrac{Q(v)}{N(v)} + \alpha\sqrt{\dfrac{2lnN(v')}{N(v)}}$ \Comment{UCT best leaf} \label{lst:line:uctheuristic}
\EndProcedure

\item[]

\Procedure{Expand}{$v,\ text$}
\For {$i = 0$, $i < length(text)$, $i{+}{+}$}  
\State $v.expand(i)$  \Comment{Create v's i-th child}
\EndFor
\State \textbf{return} $getChildren(v)$ \Comment{Return the expanded children}
\EndProcedure

\item[]

\Procedure{Simulate}{$C,\ text,\ sims,\ \network,\ \embedding,\ t$}
\State $Perturbations \gets [ \ ]$
\For {$c \in C$}
\For {$i = 0$, $i < sims$, $i{+}{+}$}
\State {$text' \gets samplePerturbation(text,\ c)$} \Comment{Ref. Figure~\ref{fig:sub_method}} \label{lst:line:sample}
\State {$x \gets \embedding(text);\ x_i' \gets \embedding(text')$} \Comment{Embed inputs}
\If {$\network(x_i') \not= \network(x)$} \label{lst:line:simulate1}  \Comment{The output class changes}
\State {$Perturbations.append(text')$}
\EndIf
\EndFor
\State {$Q(c)$} = {$max_{i\in sims}(\network_t(x)-\network_t(x'_i))$}  \Comment{Update vertex heuristic} \label{lst:line:simulate2}
\EndFor
\State \textbf{return} $Perturbations$
\EndProcedure

\item[]

\Procedure{Backpropagate}{$v,\ root$} \Comment{Propagate UCT update}
\While{$v\not=root$}
\State {$updateUCT(v)$}
\State $v \gets getParent(v)$
\EndWhile
\EndProcedure

\end{algorithmic}
\end{algorithm*}

\begin{figure}[t]
    \centering
	\includegraphics[width=\linewidth,keepaspectratio]{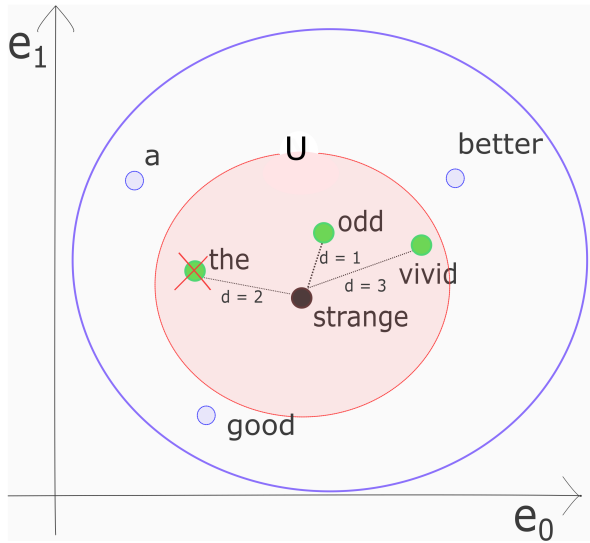}
	\caption{Substitutions are selected either randomly or according to a score calculated as a function of the distance from the original word. The sampling region (red circle) is a finite fraction of the embedding space (blue circle). Selected candidates can be filtered to enforce semantic and syntactic constraints. Word $\texttt{the}$ has been filtered out because it is not grammatically consistent with the original word $\texttt{strange}$, while words $\texttt{good}$, $\texttt{better}$ and $\texttt{a}$ are filtered out as they lie outside the neighborhood of the original word.} 
	\label{fig:sub_method}
\end{figure}

For our implementation we adopted $\textit{sims}=1000$ and $\alpha=0.5$.
 Tables~\ref{tab:mcts_table_multiple_attacks} and~\ref{tab:mcts_table} give details of MCTS experiments with single and multiple word substitutions.
\paragraph{MCTS Word Substitution Strategies} \label{app:subs}
We consider two refinements of MCTS: weighting the replacement words by importance and filtering to ensure syntactic/semantic coherence of the input text. The importance score of a word substitution is
%\MK{should this be inversely proportional, i.e. closer words are more important?}\BB{that's true, it has been changed}
inversely proportional to its distance from the original word, e.g., $pickup(w\leftarrow w')=\dfrac{1}{\abs{U}-1}(\dfrac{\sum_{u \in U \setminus \{w'\}}^{}d(w, u)}{\sum_{u \in U}^{}d(w, u)})$, where $w, w'$ are respectively the original and perturbed words, $d()$ is an $L^p$ norm of choice and $U$ a neighbourhood of $w$, whose cardinality, which must be greater than $1$, is denoted with $\abs{U}$ (as shown in Figure~\ref{fig:sub_method}). We can further filter words in the neighborhood such that only synonyms/antonyms are selected, thus guaranteeing that a word is replaced by a meaningful substitution; more details are provided in Section~\ref{sec:mcts}. While in this work we use a relatively simple method to find replacements that are syntactically coherent with the input text, more complex methods are available that try also to enforce semantic consistency~\citep{navigli2009word,ling2015two,trasksense2vec:2015}, despite this problem is known to be much harder and we reserve this for future works.

\subsection{Experimental Setup}
\begin{figure*}[t]
    \centering
	\includegraphics[width=15cm,height=15cm,keepaspectratio]{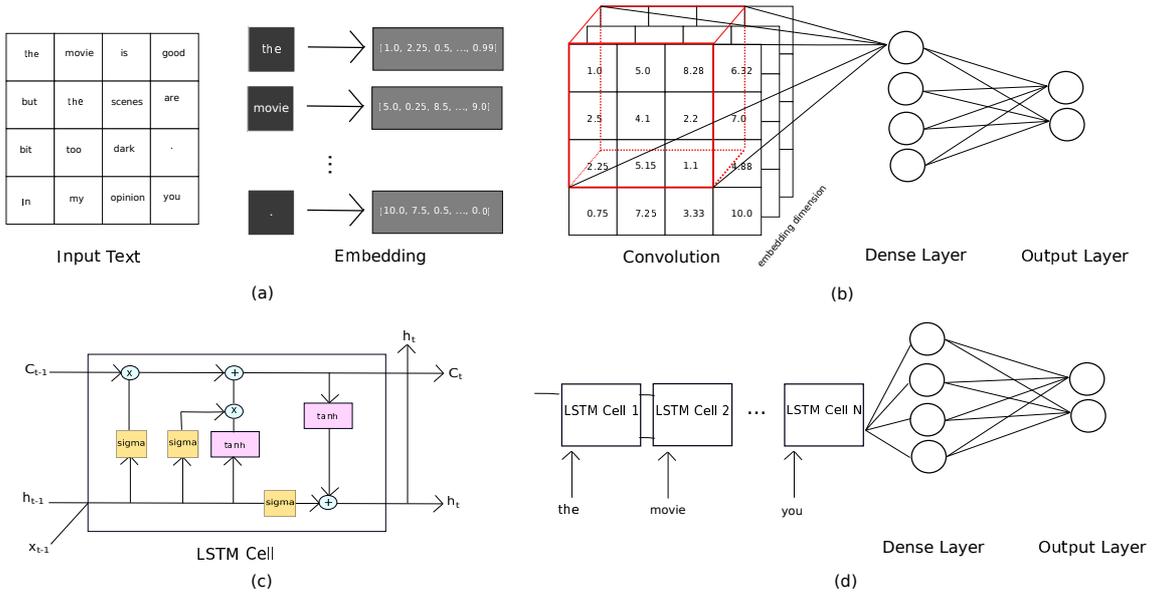}
	\caption{Architecture of CNN and LSTM \emph{vanilla} models used in this work. (a) Embedding of input words as vectors of real numbers that are passed as input to a network model that outputs the class to which a text belongs (shown here with two outputs, e.g., a \textit{positive, negative} review of a movie). (b) Convolutional network (CNN) model. (d) LSTM network model. (c) A single LSTM cell in detail.
	} 
	\label{fig:cnn_lstm}
\end{figure*}

The network architectures that have been employed in this work are shown in Figure~\ref{fig:cnn_lstm}, while the embeddings are summarised in Table~\ref{tab:embeddings}.
More details of both the embeddings and the architectures employed are provided in the main paper, Section~\ref{sec:experiments-setup}.
\subsection{Additional Robustness Results} \label{app:other-results}
\begin{figure*}[t]
    \centering
	\includegraphics[width=\linewidth,keepaspectratio]{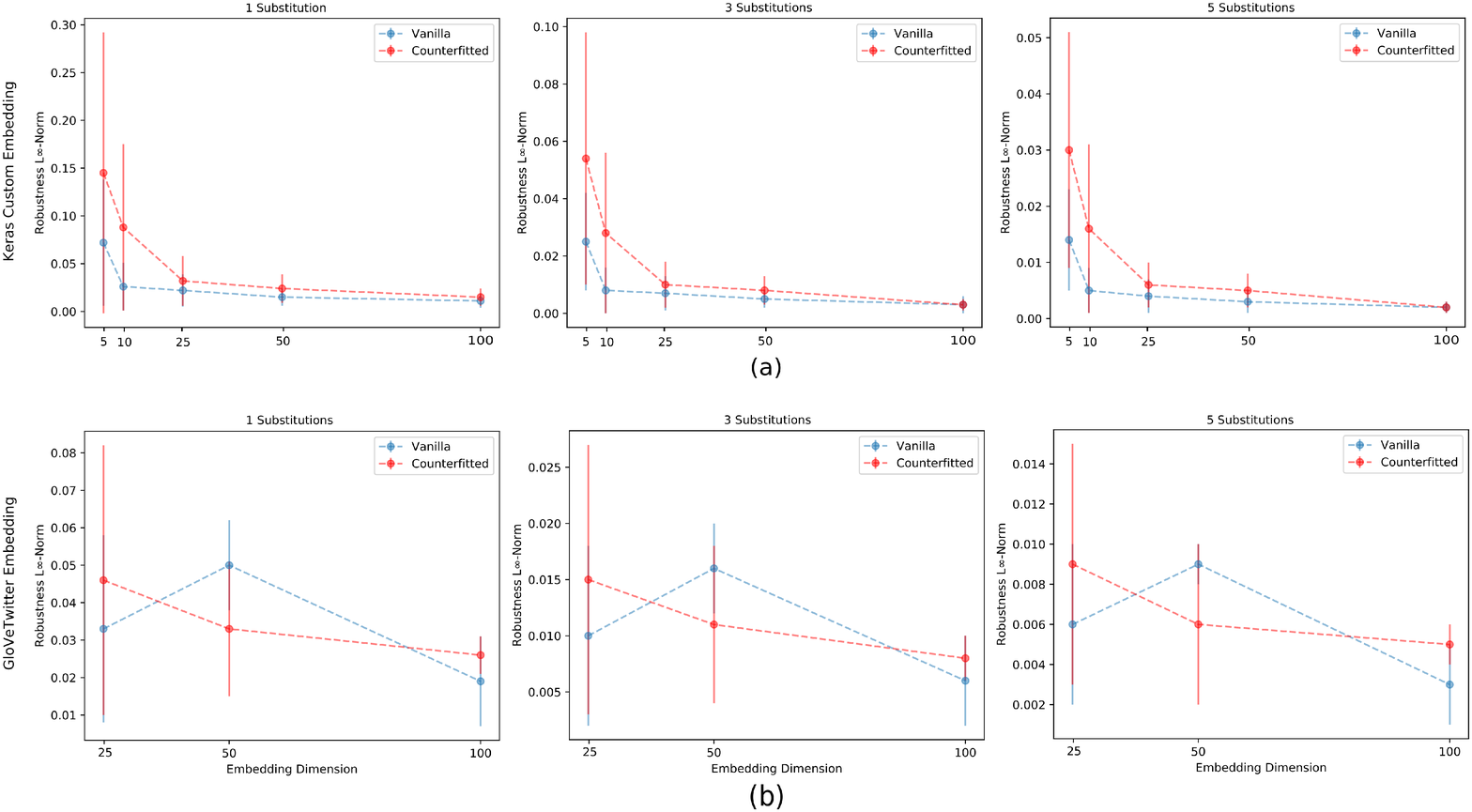}
	\caption{
	Comparison of robustness of vanilla vs. counter-fitted embeddings for an increasing number of dimensions and word substitutions on the AG News dataset. (a) Simple Keras Custom embeddings optimised for emotional polarity. (b) GloVeTwitter embeddings that encode more complex representations. Counter-fitted embeddings exhibit greater robustness on low-dimensional or simple embeddings. A reversed trend is observed on high-dimensional embeddings or more complex word representations. Values reported refer to measurements in the L$_{\infty}$-norm. } 
	\label{fig:epsreg}
\end{figure*}

In the remainder of this section we present additional experimental results of our robustness evaluation. More specifically, we show the trends of upper and lower bounds for different datasets (Tables~\ref{tab:imdb_sst_news_table},~\ref{tab:ag_table},~\ref{tab:mcts_table} and~\ref{tab:mcts_table_multiple_attacks}); include robustness results against multiple substitutions; and perform robustness comparison with counter-fitted models (Figure~\ref{fig:epsreg}). 

\begin{table*}
\centering
{Embeddings}
\\
\vspace{2 mm}
\begin{tabular}{|c|c|c|c|c|}
\hline
 \textbf{} &  \small{\textbf{DIM}} & \small{\textbf{WORDS}} &  \small{\textbf{DIAMETER}} & \small{\textbf{DIAMETER (raw)}}
 \\ \hline
  \multirow{4}{*}{\small{\textbf{Keras}}} & \small{$5$} &  \small{$177175$}  &  \small{$2.236$} &  \small{$1.144$}  \\ \cline{3-5}
                          & \small{$10$} & \small{$88587$} & \small{$3.162$} & \small{$0.957$} \\ \cline{3-5}
                          & \small{$25$} & \small{$$88587$$} & \small{$5$} & \small{$0.763$}\\ \cline{3-5}                          
                          & \small{$50$} & \small{$88587$} & \small{$7.07$} & \small{$0.664$}\\ \cline{3-5}
                          & \small{$100$} & \small{$88587$} & \small{$10$} & \small{$0.612$}\\ \hline
  \multirow{2}{*}{\small{\textbf{GloVe}}} & \small{$50$} &  \small{$400003$} &  \small{$7.071$} & \small{$10.918$} \\ \cline{3-5}
                          & \small{$100$} & \small{$400003$} & \small{$10$} & \small{$8.133$}\\ \hline
  \multirow{4}{*}{\small{\textbf{GloVeTwitter}}} & \small{$25$} &  \small{$1193517$} &  \small{$5$}& \small{$21.15$}  \\ \cline{3-5}
                          & \small{$50$} & \small{$1193517$} & \small{$7.071$} & \small{$13.947$}\\ \cline{3-5}
                          & \small{$100$} & \small{$1193517$} & \small{$10$} & \small{$13.058$}\\ \hline                          
\end{tabular}
\caption{Embeddings used for the experimental evaluation: we report the number of dimensions, the number of words in each vocabulary and the maximum distance between the two farthest words, namely the \textit{diameter}  %We report the value of the diameter
(both after normalization of the input vectors and the raw value, expressed in the L$_2$-norm). After normalization, an embedding of dimension $d$ will have a diameter equal to $\sqrt{d}$, as a consequence of scaling to $1$ the difference between maximum and minimum values for any dimension of the input. 
}
\label{tab:embeddings}
\end{table*}

\begin{table*}[t]
\centering
{IMDB}
\\
\vspace{2 mm}
\begin{tabular}{|c|c|c|c|c|}
\hline
 \textbf{} &  \small{\textbf{DIMENSION}} & \small{\textbf{ACCURACY}} &  \small{\textbf{LOWER BOUND}} \\ \hline
  \multirow{6}{*}{\textbf{Keras}} & \small{$5$} &  \small{$0.789$}  & \small{$1.358 \pm 0.604$}  \\ \cline{3-4}
                          & \small{$10$} & \small{$0.788$} & \small{$2.134 \pm 1.257$} \\ \cline{3-4}
                          & \small{$25$} & \small{$0.78$} & \small{$1.234 \pm 2.062$} \\ \cline{3-4}
                          & \small{$50$} & \small{$0.78$} & \small{$0.394 \pm 0.079$} \\ \cline{3-4}
                          & \small{$100$} & \small{$0.778$}  & \small{$0.31 \pm 0.041$} \\ 
                          \hline
  \multirow{2}{*}{\textbf{GloVe}} & \small{$50$} &  \small{$0.758$} & \small{$0.133 \pm 0.054$} \\ \cline{3-4}
                         & \small{$100$} & \small{$0.783$} & \small{$0.127 \pm 0.055$} \\ \hline
  \multirow{3}{*}{\textbf{GloVeTwitter}} & \small{$25$} &  \small{$0.739$} & \small{$0.168 \pm 0.093$} \\ \cline{3-4}
                                & \small{$50$} & \small{$0.752$} & \small{$0.143 \pm 0.02$} \\ \cline{3-4}
                                & \small{$100$} &  \small{$0.77$} & \small{$0.177 \pm 0.057$}\\ \hline
\end{tabular}
\vspace{5 mm}
\\
{Stanford Sentiment Treebank (SST)}
\\
\vspace{2 mm}
\begin{tabular}{|c|c|c|c|c|}
\hline
 \textbf{} &  \small{\textbf{DIMENSION}} & \small{\textbf{ACCURACY}} &  \small{\textbf{LOWER BOUND}} \\ \hline
  \multirow{6}{*}{\textbf{Keras}} & \small{$5$} &  \small{$0.75$}  & \small{$0.623 \pm 0.28$}  \\ \cline{3-4}
                          & \small{$10$} & \small{$0.756$} & \small{$0.449 \pm 0.283$} \\ \cline{3-4}
                          & \small{$25$} & \small{$0.757$} & \small{$0.116 \pm 0.14$} \\ \cline{3-4}
                          & \small{$50$} & \small{$0.811$} & \small{$0.029 \pm 0.012$} \\ \cline{3-4}
                          & \small{$100$} & \small{$0.818$}  & \small{$0.023 \pm 0.006$} \\ 
                          \hline
  \multirow{2}{*}{\textbf{GloVe}} & \small{$50$} &  \small{$0.824$} & \small{$0.053 \pm 0.023$} \\ \cline{3-4}
                         & \small{$100$} & \small{$0.833$} & \small{$0.028 \pm 0.015$} \\ \hline
  \multirow{3}{*}{\textbf{GloVeTwitter}} & \small{$25$} &  \small{$0.763$} & \small{$0.065 \pm 0.023$} \\ \cline{3-4}
                                & \small{$50$} & \small{$0.826$} & \small{$0.059 \pm 0.031$} \\ \cline{3-4}
                                & \small{$100$} &  \small{$0.823$} & \small{$0.0 \pm 0.0$ (NaN)}\\ \hline
\end{tabular}
\vspace{5 mm}
\\
{NEWS Dataset}
\\
\vspace{2 mm}
\begin{tabular}{|c|c|c|c|c|}
\hline
 \textbf{} &  \small{\textbf{DIMENSION}} & \small{\textbf{ACCURACY}} &  \small{\textbf{LOWER BOUND}} \\ \hline
  \multirow{2}{*}{\textbf{GloVe}} & \small{$50$} &  \small{$0.625$} & \small{$0.013 \pm 0.015$} \\ \cline{3-4}
                         & \small{$100$} & \small{$0.7$} & \small{$0.018 \pm 0.017$} \\ \hline
  \multirow{2}{*}{\textbf{GloVeTwitter}} & \small{$50$} &  \small{$0.627$} & \small{$0.009 \pm 0.006$} \\ \cline{3-4}
                                & \small{$100$} &  \small{$0.716$} & \small{$0.008 \pm 0.009$}\\ \hline
\end{tabular}
\vspace{5 mm}
\\
\caption{
Lower bound results for single-word substitutions as found by CNN-Cert and POPQORN,  respectively, on the IMDB, SST and NEWS datasets. Values reported refer to measurements in the  L$_2$-norm.}
\label{tab:imdb_sst_news_table}
\end{table*}

\begin{table*}[t]
\centering
{AG News Results: Single Word Substitution}
\\
\vspace{2 mm}
\begin{tabular}{|c|c|c|c|c|c|}
\hline
& {\textbf{\small{DIAMETER}}} & \multicolumn{2}{c|}{\textbf{\small{ACCURACY}}} & \multicolumn{2}{c|}{\textbf{\small{LOWER BOUND}}} \\
\cline{3-6}
  & {} & \textbf{\small{Vanilla}} & \textbf{\small{Counter-fitted}} & \textbf{\small{Vanilla}} & \textbf{\small{Counter-fitted}} \\ \hline
  \multirow{6}{*}{\textbf{Keras}} & \small{$5$} &  \small{$0.414$} & \small{$ 0.464$}  & \small{$0.072 \pm 0.066$} & \small{$ 0.145 \pm 0.147$}  \\ \cline{3-6}
                          & \small{$10$} & \small{$0.491$} & \small{$0.505$} & \small{$0.026 \pm 0.025$} & \small{$ 0.088 \pm 0.087$} \\ \cline{3-6}
                          & \small{$25$} & \small{$0.585$} & \small{$ 0.597$} & \small{$0.022 \pm 0.025$} & \small{$ 0.032 \pm 0.026$} \\ \cline{3-6}
                          & \small{$50$} & \small{$0.692$} & \small{$ 0.751$} & \small{$0.015 \pm 0.009$} & \small{$ 0.024 \pm 0.015$} \\ \cline{3-6}
                          & \small{$100$} & \small{$0.779$} & \small{$ 0.807$}  & \small{$0.011 \pm 0.007$} & \small{$ 0.015 \pm 0.009$} \\ 
                          \hline
  \multirow{2}{*}{\textbf{GloVe}} & \small{$50$} &  \small{$0.892$} & \small{$ 0.879$} & \small{$0.04 \pm 0.028$} & \small{$ 0.043 \pm 0.03$} \\ \cline{3-6}
                         & \small{$100$} & \small{$0.901$} & \small{$ 0.887$} & \small{$0.027 \pm 0.018$} & \small{$ 0.0 \pm 0.0$ (NaN)} \\ \hline
  \multirow{3}{*}{\textbf{GloVeTwitter}} & \small{$25$} &  \small{$0.848$} & \small{$ 0.846$} & \small{$0.033 \pm 0.025$} & \small{$ 0.046 \pm 0.036$} \\ \cline{3-6}
                                & \small{$50$} & \small{$0.877$} & \small{$ 0.866$} & \small{$0.05 \pm 0.012$} & \small{$ 0.033 \pm 0.018$} \\ \cline{3-6}
                                & \small{$100$} &  \small{$0.833$} & \small{$ 0.883$} & \small{$0.019 \pm 0.012$} & \small{$ 0.026 \pm 0.005$}\\ \hline
\end{tabular}
\vspace{2 mm}
\\
{AG News Results: Multiple Words Substitutions}
\\
\vspace{2 mm}
\begin{tabular}{|c|c|c|c|c|c|}
\hline
& {\textbf{\small{DIAMETER}}} & \multicolumn{2}{c|}{\textbf{\small{L.B. 2 SUBSTITUTIONS}}} & \multicolumn{2}{c|}{\textbf{\small{L.B. 3 SUBSTITUTIONS}}} \\
\cline{3-6}
  & {} & \textbf{\small{Vanilla}} & \textbf{\small{Counter-fitted}} & \textbf{\small{Vanilla}} & \textbf{\small{Counter-fitted}} \\ \hline
  \multirow{6}{*}{\textbf{Keras}} & \small{$5$} & \small{$0.029 \pm 0.024$} & \small{$ 0.065 \pm 0.059$}  & \small{$0.025 \pm 0.017$} & \small{$ 0.054 \pm 0.044$}  \\ \cline{3-6}
                          & \small{$10$} & \small{$0.013 \pm 0.012$} & \small{$ 0.043 \pm 0.042$} & \small{$0.008 \pm 0.008$} & \small{$ 0.028 \pm 0.028$} \\ \cline{3-6}
                          & \small{$25$} & \small{$0.011 \pm 0.008$} & \small{$ 0.015 \pm 0.012$} & \small{$0.007 \pm 0.006$} & \small{$ 0.01 \pm 0.008$} \\ \cline{3-6}
                          & \small{$50$} & \small{$0.007 \pm 0.004$} & \small{$ 0.012 \pm 0.007$} & \small{$0.005 \pm 0.003$} & \small{$ 0.008 \pm 0.005$} \\ \cline{3-6}
                          & \small{$100$} & \small{$0.006 \pm 0.004$} & \small{$ 0.006 \pm 0.004$}  & \small{$0.003 \pm 0.003$} & \small{$ 0.003 \pm 0.002$} \\ 
                          \hline
  \multirow{2}{*}{\textbf{GloVe}} & \small{$50$} &  \small{$0.02 \pm 0.013$} & \small{$ 0.02 \pm 0.014$} & \small{$0.013 \pm 0.009$} & \small{$ 0.016 \pm 0.01$} \\ \cline{3-6}
                         & \small{$100$} & \small{$0.015 \pm 0.007$} & \small{$ 0.0 \pm 0.0$ (NaN)} & \small{$0.01 \pm 0.006$} & \small{$ 0.0 \pm 0.0$ (NaN)} \\ \hline
  \multirow{3}{*}{\textbf{GloVeTwitter}} & \small{$25$} &  \small{$0.014 \pm 0.011$} & \small{$ 0.023 \pm 0.017$} & \small{$0.01 \pm 0.008$} & \small{$ 0.0015 \pm 0.012$} \\ \cline{3-6}
                                & \small{$50$} & \small{$0.024 \pm 0.005$} & \small{$ 0.015 \pm 0.009$} & \small{$0.016 \pm 0.004$} & \small{$ 0.011 \pm 0.007$} \\ \cline{3-6}
                                & \small{$100$} &  \small{$0.009 \pm 0.006$} & \small{$ 0.013 \pm 0.002$} & \small{$0.006 \pm 0.004$} & \small{$ 0.008 \pm 0.002$}\\ \hline
\end{tabular}
\\
\begin{tabular}{|c|c|c|c|c|c|}
\hline
& {\textbf{\small{DIAMETER}}} & \multicolumn{2}{c|}{\textbf{\small{L.B. 4 SUBSTITUTIONS}}} & \multicolumn{2}{c|}{\textbf{\small{L.B. 5 SUBSTITUTIONS}}} \\
\cline{3-6}
  & {} & \textbf{\small{Vanilla}} & \textbf{\small{Counter-fitted}} & \textbf{\small{Vanilla}} & \textbf{\small{Counter-fitted}} \\ \hline
  \multirow{6}{*}{\textbf{Keras}} & \small{$5$} & \small{$0.018 \pm 0.012$} & \small{$ 0.035 \pm 0.028$}  & \small{$0.014 \pm 0.009$} & \small{$ 0.03 \pm 0.021$}  \\ \cline{3-6}
                          & \small{$10$} & \small{$0.006 \pm 0.005$} & \small{$ 0.02 \pm 0.019$} & \small{$0.005 \pm 0.004$} & \small{$ 0.016 \pm 0.015$} \\ \cline{3-6}
                          & \small{$25$} & \small{$0.005 \pm 0.004$} & \small{$ 0.007 \pm 0.006$} & \small{$0.004 \pm 0.003$} & \small{$ 0.006 \pm 0.004$} \\ \cline{3-6}
                          & \small{$50$} & \small{$0.003 \pm 0.002$} & \small{$ 0.005 \pm 0.002$} & \small{$0.003 \pm 0.002$} & \small{$ 0.005 \pm 0.003$} \\ \cline{3-6}
                          & \small{$100$} & \small{$0.003 \pm 0.002$} & \small{$ 0.003 \pm 0.002$}  & \small{$0.002 \pm 0.001$} & \small{$ 0.002 \pm 0.001$} \\ 
                          \hline
  \multirow{2}{*}{\textbf{GloVe}} & \small{$50$} &  \small{$0.009 \pm 0.006$} & \small{$ 0.01 \pm 0.006$} & \small{$0.008 \pm 0.005$} & \small{$ 0.008 \pm 0.006$} \\ \cline{3-6}
                         & \small{$100$} & \small{$0.007 \pm 0.004$} & \small{$ 0.0 \pm 0.0$ (NaN)} & \small{$0.005 \pm 0.003$} & \small{$ 0.0 \pm 0.0$ (NaN)} \\ \hline
  \multirow{3}{*}{\textbf{GloVeTwitter}} & \small{$25$} &  \small{$0.007 \pm 0.005$} & \small{$ 0.011 \pm 0.008$} & \small{$0.006 \pm 0.004$} & \small{$ 0.009 \pm 0.006$} \\ \cline{3-6}
                                & \small{$50$} & \small{$0.008 \pm 0.004$} & \small{$ 0.008 \pm 0.006$} & \small{$0.009 \pm 0.001$} & \small{$ 0.006 \pm 0.004$} \\ \cline{3-6}
                                & \small{$100$} &  \small{$0.004 \pm 0.003$} & \small{$ 0.006 \pm 0.001$} & \small{$0.003 \pm 0.002$} & \small{$ 0.005 \pm 0.001$}\\ \hline
\end{tabular}
\\
\caption{Lower bound results for single (top) and multiple word (middle and bottom) substitutions, comparing vanilla and counter-fitted %and non counter-fitted
models. Robustness of counter-fitted models is superior to the vanilla  %\textit{vanilla}
counterpart, except for high-dimensional embeddings such as GloVe $100$d, where %due to the high dimensionality 
it has not been possible to obtain a bound for the counter-fitted embedding due to computational constraints  (nonetheless the counterpart lower bound is close to zero). Values reported refer to measurements in the L$_{\infty}$-norm.
}
\label{tab:ag_table}
\end{table*}

\begin{table*}[t]
\centering
{MCTS Results}
\\
\vspace{2 mm}
\begin{tabular}{|c|c|c|c|c|c|}
\hline
 \textbf{} &  \textbf{\small{EMBEDDING}} & \textbf{\small{EXEC TIME [s]}} &  \textbf{\small{SUB. (\% per-text)}} & \textbf{\small{SUB. (\% per-word)}} & \textbf{\small{UB}} \\ \hline
  \multirow{3}{*}{\textbf{\small{IMDB}}} & \small{Keras$50$d} & \small{$29.52$}  &  \small{$6.0$}  & \small{$1.4$}  &  \small{$0.41 \pm 0.04$}  \\ \cline{3-6}
  & \small{GloVe$50$d} & \small{$39.61$}  &  \small{$39.7$}  & \small{$5.1$}  &  \small{$0.39 \pm 0.016$}  \\ \cline{3-6}
  & \small{GloVeTwitter$50$d} & \small{$54.1$}  &  \small{$47.0$}  & \small{$7.7$}  &  \small{$0.329 \pm 0.015$}  \\ \hline  
  \multirow{3}{*}{\textbf{\small{AG NEWS}}} & \small{Keras$50$d} & \small{$21.09$}  &  \small{$50.0$}  & \small{$15.6$}  &  \small{$0.396 \pm 0.02$}  \\ \cline{3-6}
  & \small{GloVe$50$d} & \small{$19.25$}  &  \small{$22.4$}  & \small{$10.8$}  &  \small{$0.438 \pm 0.042$}  \\ \cline{3-6}
  & \small{GloVeTwitter$50$d} & \small{$17.75$}  &  \small{$21.4$}  & \small{$6.6$}  &  \small{$0.336 \pm 0.019$}  \\ \hline
  \multirow{3}{*}{\textbf{\small{SST}}} & \small{Keras$50$d} & \small{$8.36$}  &  \small{$52.2$}  & \small{$19.9$}  &  \small{$0.444 \pm 0.077$} \\ \cline{3-6}
  & \small{GloVe$50$d} & \small{$11.94$}  &  \small{$81.1$}  & \small{$37.4$}  &  \small{$0.385 \pm 0.024$}  \\ \cline{3-6}
  & \small{GloVeTwitter$50$d} & \small{$11.96$}  &  \small{$78.1$}  & \small{$36.3$}  &  \small{$0.329 \pm 0.024$}  \\   \hline 
  \multirow{4}{*}{\textbf{\small{NEWS}}} & \small{GloVe$50$d} & \small{$75.76$}  &  \small{$96.5$}  & \small{$34.0$}  &  \small{$0.405 \pm 0.045$}  \\ \cline{3-6}
  & \small{GloVe$100$d} & \small{$79.31$}  &  \small{$89.7$}  & \small{$29.1$}  &  \small{$0.442 \pm 0.042$}  \\ \cline{3-6}
  & \small{GloVeTwitter$50$d} & \small{$77.74$}  &  \small{$90.9$}  & \small{$30.6$}  &  \small{$0.314 \pm 0.033$}  \\ \cline{3-6}  
  & \small{GloVeTwitter$100$d} & \small{$81.29$}  &  \small{$89.7$}  & \small{$27.7$}  &  \small{$0.417 \pm 0.042$}  \\  \hline                    
\end{tabular}
\\
\caption{Upper bound results for single-word substitutions as found by MCTS. 
We report: the average execution time for each experiment; the percentage of texts for which we have found at least one successful single-word substitution (which results in a class change) and 
the approximate ratio that selecting randomly 1 word from a text we find a replacement that is successful; the distance to the closest meaningful perturbation %that is the closest 
to the original word found, namely an upper bound (differently from Table~\ref{tab:msr_gap_table} and for completeness, here values are reported only considering the values for those words where the perturbations were successful). 
Values reported refer to measurements in the  L$_2$-norm.}
\label{tab:mcts_table}
\end{table*}

\begin{table*}[t]
\centering
{MCTS Multiple Substitutions}
\\
\vspace{2 mm}
\begin{tabular}{|c|c|c|c|c|c|c|c|}
\hline
& {\textbf{\small{EMBEDDING}}} & \multicolumn{2}{c|}{\textbf{\small{2 SUBSTITUTIONS}}} & \multicolumn{2}{c|}{\textbf{\small{3 SUBSTITUTIONS}}} & \multicolumn{2}{c|}{\textbf{\small{4 SUBSTITUTIONS}}} \\
\cline{3-8}
  & {} & \textbf{\small{\% per-text}} & \textbf{\small{\% per-word}} & \textbf{\small{\% per-text}} & \textbf{\small{\% per-word}} & \textbf{\small{\% per-text}} & \textbf{\small{\% per-word}} \\ \hline
  \multirow{3}{*}{\textbf{\small{IMDB}}} & \small{Keras$50$d} & \small{$8.5$}  & \small{$ 5.0$}  &  \small{$13.4$}  & \small{$ 5.9$}  & \small{$18.2$}  & \small{$ 6.6$}  \\ \cline{3-8}
  & \small{GloVe$50$d} & \small{$43.8$}  & \small{$ 17.7$}  &  \small{$52.0$}  & \small{$ 21.6$}  & \small{$57.5$}  & \small{$ 24.5$}  \\ \cline{3-8}
  & \small{GloVeTwitter$50$d} & \small{$44.1$}  & \small{$ 18.3$}  &  \small{$49.3$}  & \small{$ 23.0$}  & \small{$57.1$}  & \small{$ 26.4$}  \\ \hline  
  \multirow{3}{*}{\textbf{\small{AG NEWS}}} & \small{Keras$50$d} & \small{$68.1$}  & \small{$ 27.5$}  &  \small{$72.7$}  & \small{$ 38.3$}  & \small{$83.3$}  & \small{$ 47.9$}  \\ \cline{3-8}
  & \small{GloVe$50$d} & \small{$31.4$}  & \small{$ 15.8$}  &  \small{$33.7$}  & \small{$ 16.8$}  & \small{$37.0$}  & \small{$ 19.7$}  \\ \cline{3-8}
  & \small{GloVeTwitter$50$d} & \small{$23.8$}  & \small{$ 12.5$}  &  \small{$23.8$}  & \small{$ 15.3$}  & \small{$38.0$}  & \small{$ 18.4$}  \\ \hline
  \multirow{3}{*}{\textbf{\small{SST}}} & \small{Keras$50$d} & \small{$64.8$}  & \small{$ 33.0$}  &  \small{$74.7$}  & \small{$ 40.2$}  & \small{$78.0$}  & \small{$ 48.7$} \\ \cline{3-8}
  & \small{GloVe$50$d} & \small{$89.4$}  & \small{$ 58.0$}  &  \small{$96.4$}  & \small{$ 70.8$}  & \small{$97.6$}  & \small{$ 76.5$}  \\ \cline{3-8}
  & \small{GloVeTwitter$50$d} & \small{$88.3$}  & \small{$ 57.8$}  &  \small{$94.1$}  & \small{$ 69.1$}  & \small{$95.3$}  & \small{$ 74.9$} \\   \hline 
  \multirow{4}{*}{\textbf{\small{NEWS}}} & \small{GloVe$50$d} & \small{$98.8$}  & \small{$ 55.4$}  &  \small{$97.3$}  & \small{$ 62.5$}  & \small{$97.3$}  & \small{$ 68.6$}  \\ \cline{3-8}
  & \small{GloVe$100$d} & \small{$100.0$}  & \small{$ 46.8$}  &  \small{$95.0$}  & \small{$ 68.0$}  & \small{$96.0$}  & \small{$ 65.2$}  \\ \cline{3-8}
  & \small{GloVeTwitter$50$d} & \small{$94.5$}  & \small{$ 50.5$}  &  \small{$97.5$}  & \small{$ 63.0$}  & \small{$97.5$}  & \small{$ 71.9$}  \\ \cline{3-8}  
  & \small{GloVeTwitter$100$d} & \small{$92.7$}  & \small{$ 49.9$}  &  \small{$98.1$}  & \small{$ 58.2$}  & \small{$98.3$}  & \small{$ 65.3$} \\  \hline                    
\end{tabular}
\\
\caption{Upper bound results for multiple-word substitutions as found by MCTS. We report the percentage of texts for which we have found at least a single-word substitution and the approximate ratio that selecting randomly $k$ words from a text (where $k$ is the number of substitutions allowed) we find a replacement that is successful. We do not report the average execution times as they are (roughly) the same as in Table \ref{tab:mcts_table}. Values reported refer to measurements in the L$_2$-norm. For more than $1$ substitution, values reported are an estimate on several random replacements, as it quickly becomes prohibitive to cover all the possible multiple-word combinations.}
\label{tab:mcts_table_multiple_attacks}
\end{table*}

\end{document}